\documentclass[10pt, a4paper]{article}

\usepackage[final]{lrec2026} 
\usepackage{xurl}
\usepackage{graphicx}
\usepackage{subcaption}
\usepackage{placeins}   

\title{Mapping Discourse Reframing: A Multi-Layer Network Approach to Italian HPV Vaccine Discourse on X (2010-2024)}

\name{Lorella Viola} 

\address{Vrije Universiteit Amsterdam\\
         De Boelelaan 1105, 1081 HV Amsterdam \\
         l.viola@vu.nl\\}
         
\abstract{Understanding how online narratives travel through coalitions is critical for identifying information disorder, yet computational analyses often rely on conservative network constructions that erase initially sparse but salient signals. This paper proposes a novel multi-layer framework that captures low-frequency signals of emerging information disorder allowing for locating where online discourse is reframed and amplified over time. The use case is 14 years of Italian discourse on X regarding the Human Papillomavirus (HPV) vaccine across three pivotal epochs (2010–2024). Utilizing hashtag co-occurrence networks, we introduce a dual-layer approach. We first identify robust core discourse coalitions through conservative community detection, revealing a stable prevention-oriented backbone contrasted with increasingly separable skepticism coalitions. We then introduce a `coverage' layer and project fringe hashtags into core coalitions based on weighted connectivity. Using a manually labelled set of skeptical and conspiratorial seed tweets, we demonstrate that this core–coverage projection significantly improves the recovery of long-tail, problematic hashtags while preserving an interpretable coalition structure. Our findings characterize the structural maturation of polarized narratives and provide a methodology for mapping how discourse is reframed and amplified by information disorder over time.
 \\ \newline  }

\begin{document}

\maketitleabstract
\begin{center}
\small
\textit{To appear in the Proceedings of the LREC2026 Information Disorder (InDor) Workshop. 
The proceedings version will be licensed under CC BY-NC 4.0.}
\end{center}

\section{Introduction}
Discourses rise and fall in popularity not arbitrarily, but because they align with the prevailing social, political, and economic contexts in which they are created and perpetuated \citep{robertson_shifting_1998}. For example, public conceptualisations and discussions of health reflect more than just health: they also offer privileged viewpoints to understand how ideas, beliefs and explanations for what health is and what determines it change over time, for example, in response to increased scrutiny or societal shifts. In turn, understanding the structural and temporal evolution of health discourse is essential to identify mechanisms of information disorder and inform democratic countermeasures. This is particularly important where harms include public health risks such as the erosion of trust in medical recommendations. 

Especially in digital ecosystems, misleading, conspiratorial, or polarizing claims rarely propagate in isolation \citep{viola_barren_2025,carpiano_confronting_2023,durmaz_dramatic_2022,quintana_polarization_2022,gunaratne_temporal_2019}; instead, they travel through narrative coalitions that infiltrate core public health narratives and reframe topics, institutions, and responsibilities. Despite the exponential growth in scholarship mapping the impact and content of disinformation \citep{wiggins_nothing_2023,bonnevie_quantifying_2021,chen_effects_2021,stabile_sex_2019,van_prooijen_conspiracy_2018}, a significant computational challenge remains: information disorder signals often reside in the `long tail' of discourse, that is a large number of niche or short-lived elements which in a communication ecosystem individually appear only a few times, but collectively make up a substantial share of discourse diversity \citep{kordumova_exploring_2016}. Because emerging disinformation or conspiratorial narratives may initially present as sparse, low-frequency signals, traditional network analyses that rely on conservative filtering (e.g., high edge-count thresholds) risk erasing the very salient signals that characterize the early stages of narrative reframing. This erasure limits our ability to observe how fringe skepticism matures into robust, structurally separable coalitions that can eventually contest the prevention-oriented backbone of public health.

This paper addresses this gap by proposing a novel multi-layer framework designed to capture low-frequency signals without sacrificing the interpretability of the core discourse. The use case is 14 years of Italian discourse on X (formerly Twitter) surrounding the human papillomavirus (HPV) vaccine. This vaccine has historically been a lightning rod for anti-vaccination campaigns and gender-based disinformation \citep{viola_barren_2025,kornides_exploring_2023,calo_misinformation_2021} offering a unique longitudinal lens into how discourse reframing undermines public trust. Specifically, this study maps the structural maturation of narrative coalitions as manifested through hashtags in Italian posts about HPV across three pivotal epochs (2010–2014, 2015–2019, 2020–2024). The assumption is that hashtags can be interpreted as cultural products \citep{la_rocca_research_2022,zappavigna_twitter_2016,weller_twitter_2013} and that these cultural products act as markers of conceptual associations and communal identity. Using hashtag co-occurrence networks, we introduce a dual-layer approach: a `core' layer defined by conservative community detection ($k$-core $\geq 2$) to identify the stable thematic backbone, and a `coverage' layer (edge count $\geq 1$) that captures the long tail. By projecting these fringe hashtags into the core coalitions based on weighted connectivity, we provide a methodology for locating where and how health narratives are reframed and amplified over time. Using a manually labelled set of skeptical and conspiratorial seed posts, we demonstrate that this projection significantly improves the recovery of problematic hashtags that hijack health narratives and that would otherwise be lost in traditional network constructions.

\section{Previous studies}
\label{sec:lit. review}
Research on health discourse has long emphasized its temporal dimension, showing how shifts in public sentiment and dominant narratives can shape, sometimes rapidly, policy agendas and institutional responses \citep{mcphail_fat_2015,berridge_marketing_2007,lawrence_framing_2004}. This work highlights the dynamic interplay between public priorities, political decision-making, and the framing of responsibility and risk. A well-known example is the case of the HPV vaccine in Japan, when following a surge in media reports of unconfirmed adverse events, often framed through conspiracy-adjacent narratives of government and pharmaceutical collusion, in 2013 the Japanese Ministry of Health, Labour and Welfare (MHLW) suspended its proactive recommendation for the vaccine \citep{yagi_hpv_2024}. This suspension lasted nearly nine years, ending only in April 2022 and it is frequently cited in academic literature as a cautionary tale of how information disorder and safety concerns can lead to the total collapse of a public health program \citep{larson_volatility_2021}. 

Studies foreground how health meanings evolve over time and why such evolution matters for understanding both governance and public contestation around health. With specific reference to vaccine discourse, recent work has examined the growth and transformation of anti-vaccine activism and the changing structure of online vaccine discussion \citep{britt_evolution_2023,carpiano_confronting_2023,crupi_echoes_2022,durmaz_dramatic_2022,monsted_characterizing_2022,quintana_polarization_2022}. Across studies, several recurring patterns have emerged: increasing attention to vaccination topics online; highly polarised interaction dynamics in which users preferentially engage with like-minded others; and a growing alignment between vaccine skepticism and political identity, including far-right affiliation. Importantly, research has shown that anti-vaccination groups, though smaller than pro-vaccination communities, can account for a disproportionate share of content and remain highly active \citep{gunaratne_temporal_2019}. These dynamics are often mediated by trust: information tends to be accepted not primarily on accuracy grounds, but on whether it circulates within trusted in-groups \citep{quintana_polarization_2022}.
This important literature, however, has often addressed vaccine discourse at a broad level (without focusing on a specific vaccine) or concentrates primarily on COVID-19 vaccination. Moreover, English-language studies substantially outnumber those in other languages, limiting our understanding of how vaccination narratives evolve in different sociolinguistic and political contexts. This motivates further work on non-English case studies and on vaccine-specific discourses, such as the Italian online debate around the HPV vaccine.

In terms of online cultural artifacts, scholarship on hashtags has expanded significantly over the past two decades \citep{la_rocca_research_2022}. While hashtags originated as indexing tools for grouping content, research shows that they have evolved into multi-functional instruments that operate simultaneously as thematic markers, navigation devices, and forms of self-expression (e.g., \#metoo). Because hashtags combine descriptive signals with affective and communicative meanings, they can be studied through linguistic, cultural, and media perspectives. A growing body of work argues that hashtags operate as speech acts and cultural objects: they acquire meaning through collective use and are continually redefined via interaction. From this perspective, hashtags are not merely employed by users but are also transformed by them, contributing to the construction of social reality \citep{budnik_dynamic_2019}. This approach has been particularly influential in studies of hashtag activism, where hashtags function as rallying points for collective action and solidarity, enabling mobilisation and identity formation around political and social causes \citep{dobrin_hashtag_2020,ross_discursive_2020,sebeelo_hashtag_2021}. In these contexts, hashtags can become condensed manifestos of movements and potent markers of affiliation \citep{la_rocca_research_2022}. 

This scholarship supports the view of hashtags as cultural symbols carrying semantic content that is continuously renegotiated. This conceptualisation motivates the present study’s focus on longitudinal changes in HPV-related hashtag use in Italian posts as a way to understand evolving health narratives and their entanglement with broader social dynamics.

\section{The HPV vaccine}
HPV, the most prevalent sexually transmitted infection globally, affects both men and women, with an estimated 80\%-90\% of individuals contracting it during their lifetime \citep{kombe_kombe_epidemiology_2021}. Of the 100 types of HPV, 13 may induce cancer, with cervical cancer being the most common HPV-related cancer and the fourth most common cancer among women worldwide \citep{who_cancer_2022}. In terms of public reception, since its approval by the Food and Drugs Administration (FDA) and the European Medical Agency (EMA) in 2006, Gardasil, the HPV vaccine licensed by Merck in the same year, has faced multiple anti-vaccination campaigns around the world. Concerns have included fears of promoting sexual promiscuity, providing false security, mandatory vaccination issues, access disparities, unreported adverse effects, and corruption \citep{chen_effects_2021,briones_when_2012,vamos_hpv_2008}. These concerns have been amplified through media like Andi Reiss’s 2018 documentary `Sacrificial Virgins', which sparked controversy and government pressure in Australia \citep{gwynne_sacrificial_2020}, works like `The HPV Vaccine On Trial' \citep{holland_hpv_2018} and retracted articles on fertility issues \citep{delong_retracted_2018}.

In addition to the already cited Japanese case, disinformation and conspiracy theories, typically revolving around recurring tropes, have also been identified in the discourse of HPV anti-vaccination campaigns \citep{smith_infertility_2024,khalil_human_2023,wakefield_infertility_2022}. In China, some conspiracists claim that the HPV vaccine is both a profit-driven scheme by the government and a biological weapon from western nations aimed at eradicating the Chinese ethnic group \citep{chen_effects_2021}. Broader global narratives include accusations that governments and pharmaceutical companies fabricate data on vaccine efficacy and safety \citep{khalil_human_2023}, allegations that the vaccine causes infertility and primary ovarian insufficiency (POI), depopulation conspiracy theories \citep{viola_barren_2025,smith_infertility_2024}, and claims that vaccinated individuals shed spike proteins, causing illness in others \citep{wakefield_infertility_2022}. 

To complicate the information space even more, some theories have been substantiated, such as the 2017 Italian exposé on corruption during the Gardasil approval process \citep{borella_report_2017,gabanelli_se_2016}, thus making 
Italy a particularly valuable use case for the present study. Italy was also the first European country to adopt a population-based HPV vaccination strategy for 12-year-old girls \citep{mennini_hpv_2022,gabutti_human_2021}. Additional age groups were proposed as secondary targets, including 25-year-old women already involved in HPV screening services and a potential third cohort of women between 12 and 25 years old. The 2017–2019 National Immunisation Plan (NIP) expanded to include both sexes as primary targets for HPV vaccination during adolescence, preferably before sexual debut \citep{mennini_hpv_2022}. Currently, the vaccine is offered for free; it is not mandatory but recommended for boys and girls between 11 and 15 years of age, with some regions extending the gratuity up until 26 years of age.

\section{Data and methodology}
Data retrieval was conducted through targeted queries that extracted posts from X containing specific hashtags, including \#HPV, \#Gardasil, \#papillomavirus, \#papilloma.\footnote{Data collection was performed in 2025 on the Apify platform by executing a Twitter/X scraping Actor (Actor ID: nfp1fpt5gUlBwPcor) via Apify Actor runs (Apify API v2 / Apify Console), and exporting the resulting dataset from the Actor run output.} Even though posts were collected with the platform language parameter set to Italian (lang=it), an additional post-hoc language filter was necessary because the platform’s language metadata is not fully reliable, mostly due to language tags produced by automated classifiers. Language identification was performed using fastText’s pre-trained language identification model (lid.176), and only tweets predicted as Italian (it) were kept for downstream analysis. The resulting corpus spans from 2 January 2010 to 30 December 2024 ($\approx$5,476 days, $\approx15$ years) and contains 4,895 unique post records from 2,252 unique usernames (1,910 unique user IDs). The 2,394 hashtags are present in 904 posts (18.47\%), comprising 754 unique hashtags. The dataset also includes several attributes such as the post texts, likes, replies, reposts, shares, and quotes count. A detailed description is given in Table \ref{tab:dataset-stats} in the \hyperref[appendix]{Appendix}. The dataset was finally pseudonymised and can be provided upon request to the author.

To map long-term changes in Italian HPV vaccine discourse, the study models hashtags as markers of discourse by analysing their co-occurrence structure over time.
As already discussed, the methodology was designed to overcome the loss of early information disorder signals inherent in longitudinal social media analysis whereas traditional network approaches are often applied with aggressive filtering to ensure structural interpretability. To recover such long-tail signals and identify where information disorder typically originates, we implemented a dual-layer hashtag co-occurrence framework that maintains the rigour of conservative community detection while capturing the emerging narrative reframing found in lower-frequency discourse. 

Based on their timestamp, posts were first segmented into three pivotal epochs: 2010–2014 (initial policy implementation), 2015–2019 (policy Expansion), and 2020–2024 (COVID and Post-COVID). Following the research design (i.e., hashtags as cultural products that signal conceptual association and communal alignment) all hashtags were extracted. For each post and year,  we then constructed an undirected weighted co-occurrence table of hashtag pairs (bigrams), where an edge between hashtags $h_i$ and $h_j$ is created whenever they appeared in the same post and the edge weight equalled the number of posts in that year in which the pair co-occurs. Yearly edges were aggregated within each epoch to obtain an epoch-specific weighted hashtag co-occurrence graph. Table \ref{tab:hashtags-bigrams-epoch} illustrates the distribution of posts, hashtags and hashtags bigrams across epochs. 

\begin{table}[t]
\centering
\small
\setlength{\tabcolsep}{6pt}
\begin{tabular}{l r r r}
\hline
\textbf{Epoch} & \textbf{Posts} & \textbf{Hashtags} & \textbf{Bigrams} \\
\hline
2010--2014 & 1{,}277 & 640  & 812 \\
2015--2019 & 1{,}514 & 1{,}095 & 1{,}932 \\
2020--2024 & 2{,}104 & 659  & 1{,}127 \\
\hline
\end{tabular}
\caption{Post, hashtag and hashtag-bigram counts per epoch.}
\label{tab:hashtags-bigrams-epoch}
\end{table}

From the distribution of hashtags and hashtags bigrams across epochs, interesting observations can already be drawn. There is a clear surge in both unique hashtags and bigram connections during the 2015–2019 epoch. This aligns with the `Policy Expansion' phase, suggesting that the discourse became more dense and diverse. Moreover, in all epochs, the number of bigrams significantly exceeds the number of hashtags, indicating a highly interconnected network where hashtags are rarely used in isolation.

To identify the stable thematic backbone of the discourse, we first construct a core layer. In this layer, we apply conservative network constraints to filter out noise and ensure the robustness of the identified coalitions, specifically, only edges with a weight $\geq 2$ (co-occurrence in at least two unique posts) are retained. To ensure that every hashtag is part of a conversation circle rather than just a lonely tag attached to a single main topic, we also apply $k$-core constraint $\geq 2$, pruning all nodes that are not part of a subgraph where every node is connected to at least two others. 

Community structure within each epoch graph was identified using a hierarchical Girvan-Newman community detection procedure \citep{girvan_community_2002}; this algorithm reflects the number of shortest paths that pass through an edge. By progressively removing edges with the highest betweenness, it effectively disassembles the network along its most significant connections, revealing peripheral community structures within the graph. This method is especially suited to smaller or moderately-sized networks where the clarity of the hierarchical structure is paramount. Using this algorithm, we detected the core coalitions, that is the most stable discourse. To enable diachronic comparison, communities were matched across epochs using Jaccard overlap of their core hashtag sets, producing stable community identifiers.

We then computed the coverage layer. This layer is meant to capture the information disorder signals, which are often sparse and initially disconnected from the mainstream, as already discussed. This layer retains every hashtag co-occurrence (edge weight $\geq 1$) and removes the $k$-core constraint. This inclusive approach ensures that low-frequency hashtags such as emerging skepticism markers are captured rather than discarded. Finally, we implement a core–coverage projection. Rather than treating the two layers as separate entities, we project the coverage hashtags into the existing core coalitions based on their weighted relative connectivity. Specifically, a hashtag from the coverage layer was assigned to a core coalition if it shared an edge with at least one node in that core coalition and its strongest weighted connection (or the sum of its connections) gravitates towards that specific cluster. Hashtags that shared no edges with any core nodes remained unassigned. This method allowed us to observe how fringe, long-tail narratives attempt to infiltrate or reframe established core narratives, for example showing how a conspiracy hashtag might attach itself to a core prevention node. 

Finally, to validate the effectiveness of this projection, we manually labelled a set of seed posts containing verified skeptical or conspiratorial content. Seed posts were selected via Critical Discourse Analysis (CDA)-guided manual identification of canonical vaccine-controversy frames as found in the literature (e.g., safety harms, corruption/pharma capture, freedom/control, depopulation rhetoric, institutional distrust). The author then manually flagged posts in the corpus that clearly instantiated these frames and used the hashtags appearing in these posts to build a small seed lexicon for weak supervision. By measuring the recovery rate of these seeds within the projected layer versus a standard single-layer $k$-core graph, we demonstrate that the Core–Coverage framework captures significantly more salient information disorder signals while preserving an interpretable structural map of the discourse evolution.

\section{Analysis and Results} 
First, we report the size of the \emph{largest connected component} (LCC), i.e., the number of nodes in the largest subgraph in which all nodes are mutually reachable via paths, together with the total number of disconnected components in the network (implicitly). Coverage networks are larger and typically include more peripheral components, while the core networks concentrate mass into fewer, denser structures. Second, we report the \emph{within-community edge share}, defined as the fraction of edges whose endpoints fall within the same detected community. As expected, this share is substantially higher in the core networks (2010--2014: 0.90 vs.\ 0.76; 2015--2019: 0.95 vs.\ 0.79), indicating that sparsification produces a clearer modular backbone, whereas the coverage layer preserves more cross-community bridges and niche connections that increase between-community linking. The results are displayed in Table \ref{tab:core-coverage-diagnostics-all}.

\begin{table}[t]
\centering
\footnotesize
\setlength{\tabcolsep}{3.5pt}
\begin{tabular}{l l r r r r r}
\hline
\textbf{Epoch} & \textbf{Layer} & \textbf{N} & \textbf{E} & \textbf{C} & \textbf{LCC} & \textbf{W-in} \\
\hline
2010--2014 & Core     & 46  & 99   & 3  & 40  & 0.90 \\
2010--2014 & Coverage & 150 & 403  & 5  & 137 & 0.76 \\
2015--2019 & Core     & 86  & 224  & 7  & 69  & 0.95 \\
2015--2019 & Coverage & 349 & 1{,}169 & 16 & 312 & 0.79 \\
2020--2024 & Core     & 40  & 84   & 7  & 24  & 1.00 \\
2020--2024 & Coverage & 268 & 858  & 17 & 220 & 0.84 \\
\hline
\end{tabular}
\caption{Core vs.\ coverage diagnostics across epochs. N = nodes; E = edges; C = connected components; LCC = size of the largest connected component; W-in = within-community edge share.}
\label{tab:core-coverage-diagnostics-all}
\end{table}

\begin{figure*}[!t]
  \centering
  \begin{subfigure}[t]{0.49\textwidth}
    \centering
    \includegraphics[width=\linewidth]{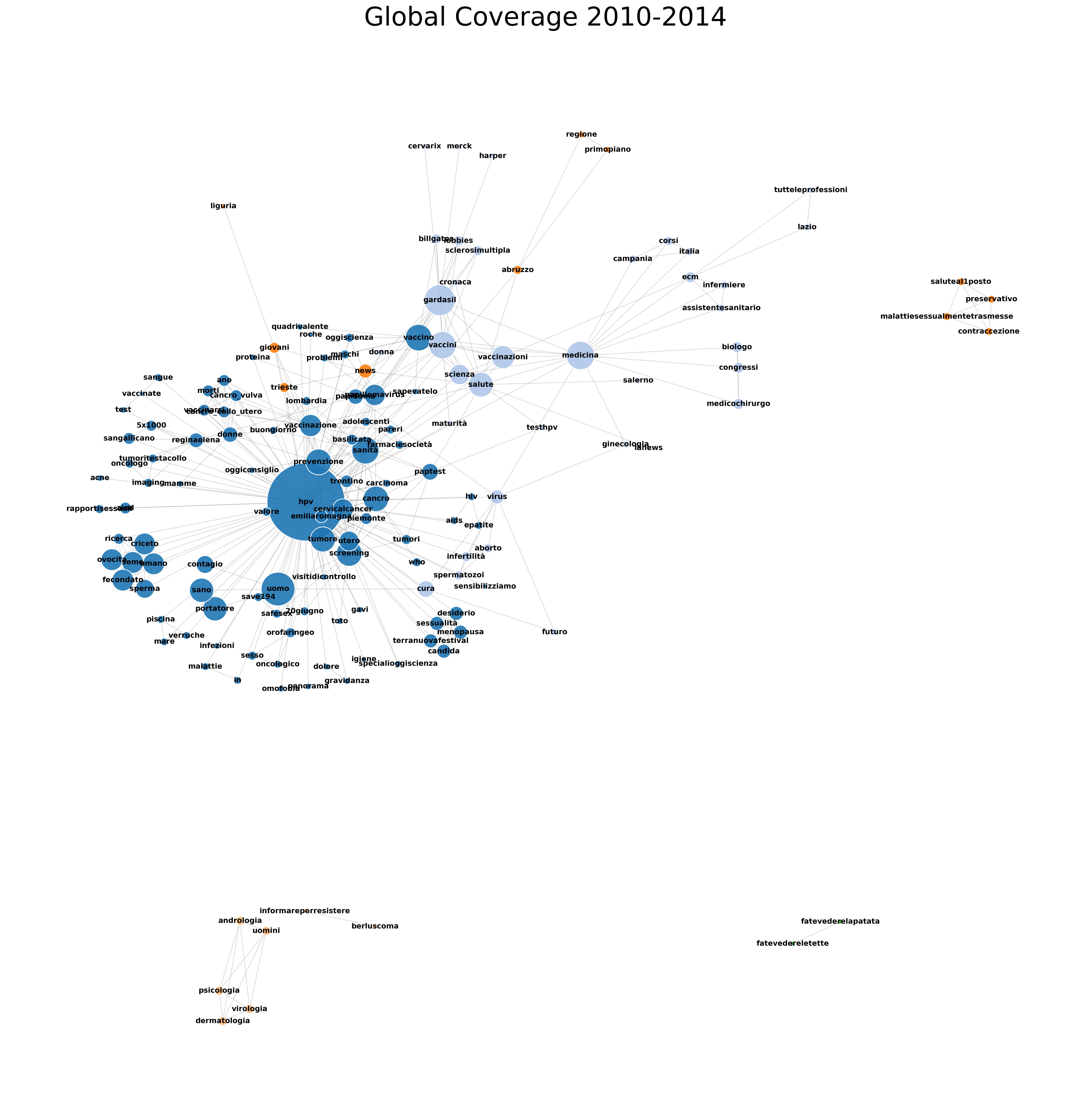}
    \caption{Coverage network (2010--2014).}
  \end{subfigure}\hfill
  \begin{subfigure}[t]{0.49\textwidth}
    \centering
    \includegraphics[width=\linewidth]{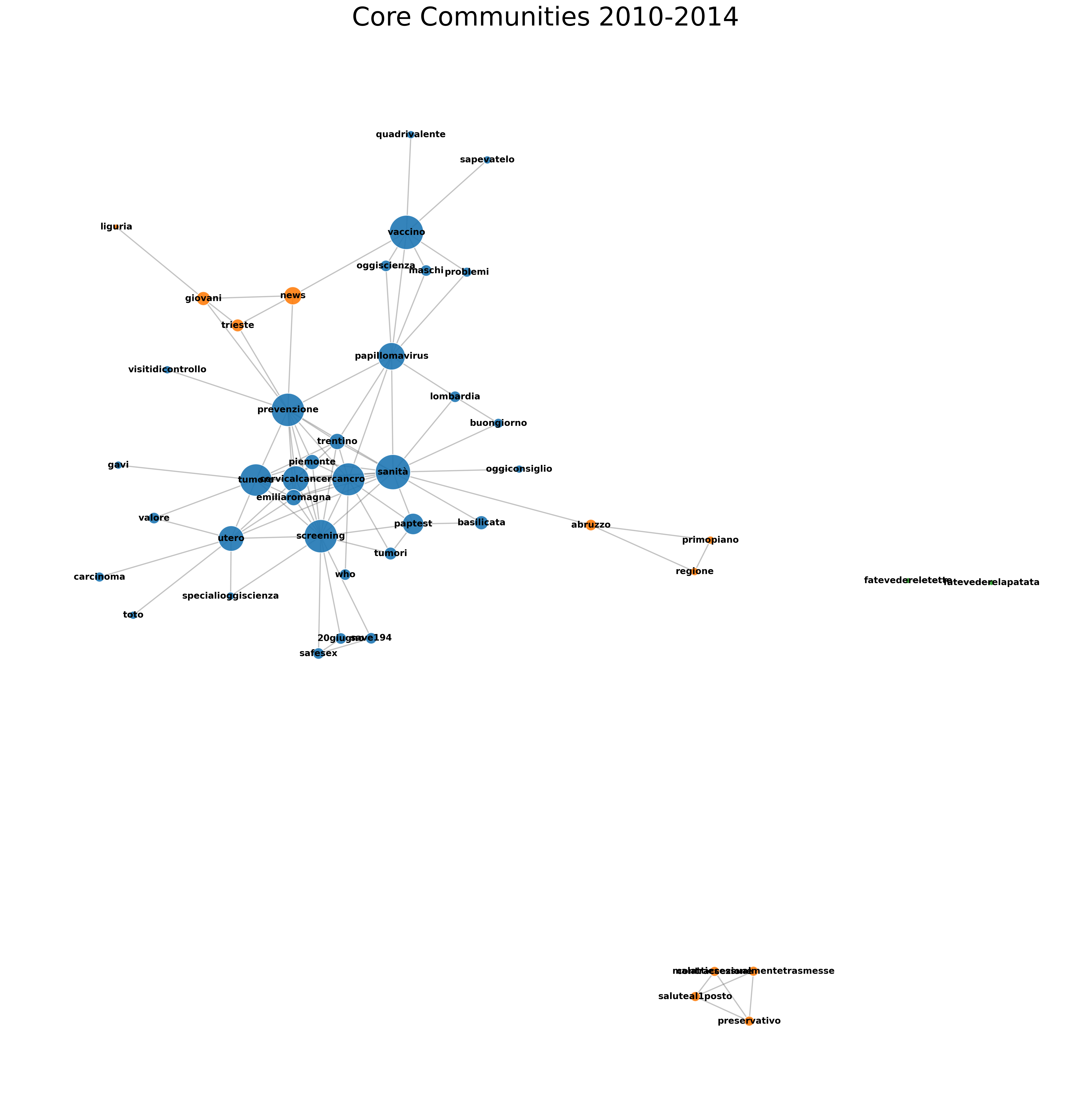}
    \caption{Core network (2010--2014).}
  \end{subfigure}
  \caption{Hashtag co-occurrence networks for 2010--2014 (coverage vs.\ core).}
  \label{fig:networks-2010-2014}
\end{figure*}

\begin{figure*}[!t]
  \centering
  \begin{subfigure}[t]{0.49\textwidth}
    \centering
    \includegraphics[width=\linewidth]{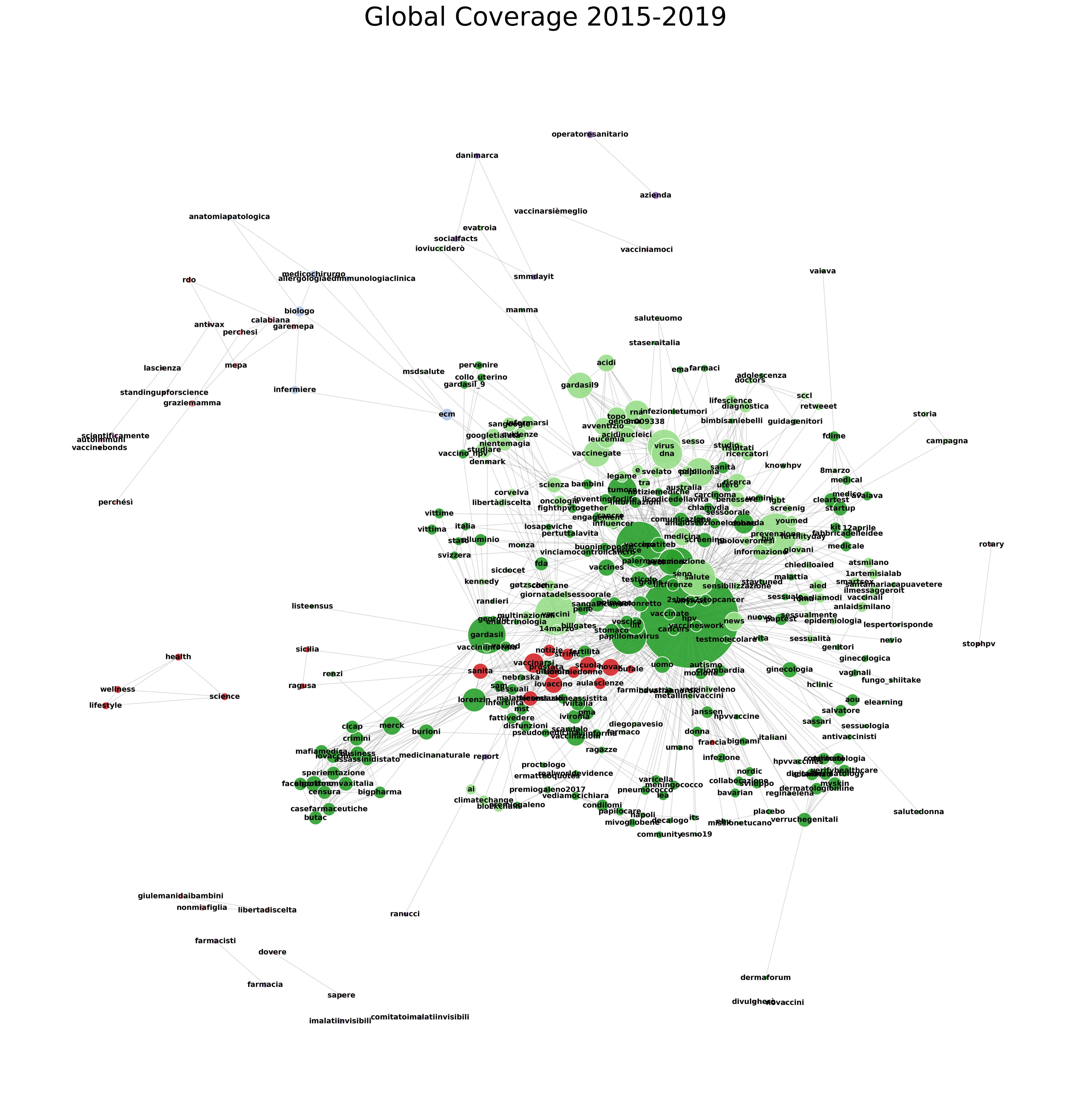}
    \caption{Coverage network (2015--2019).}
  \end{subfigure}\hfill
  \begin{subfigure}[t]{0.49\textwidth}
    \centering
    \includegraphics[width=\linewidth]{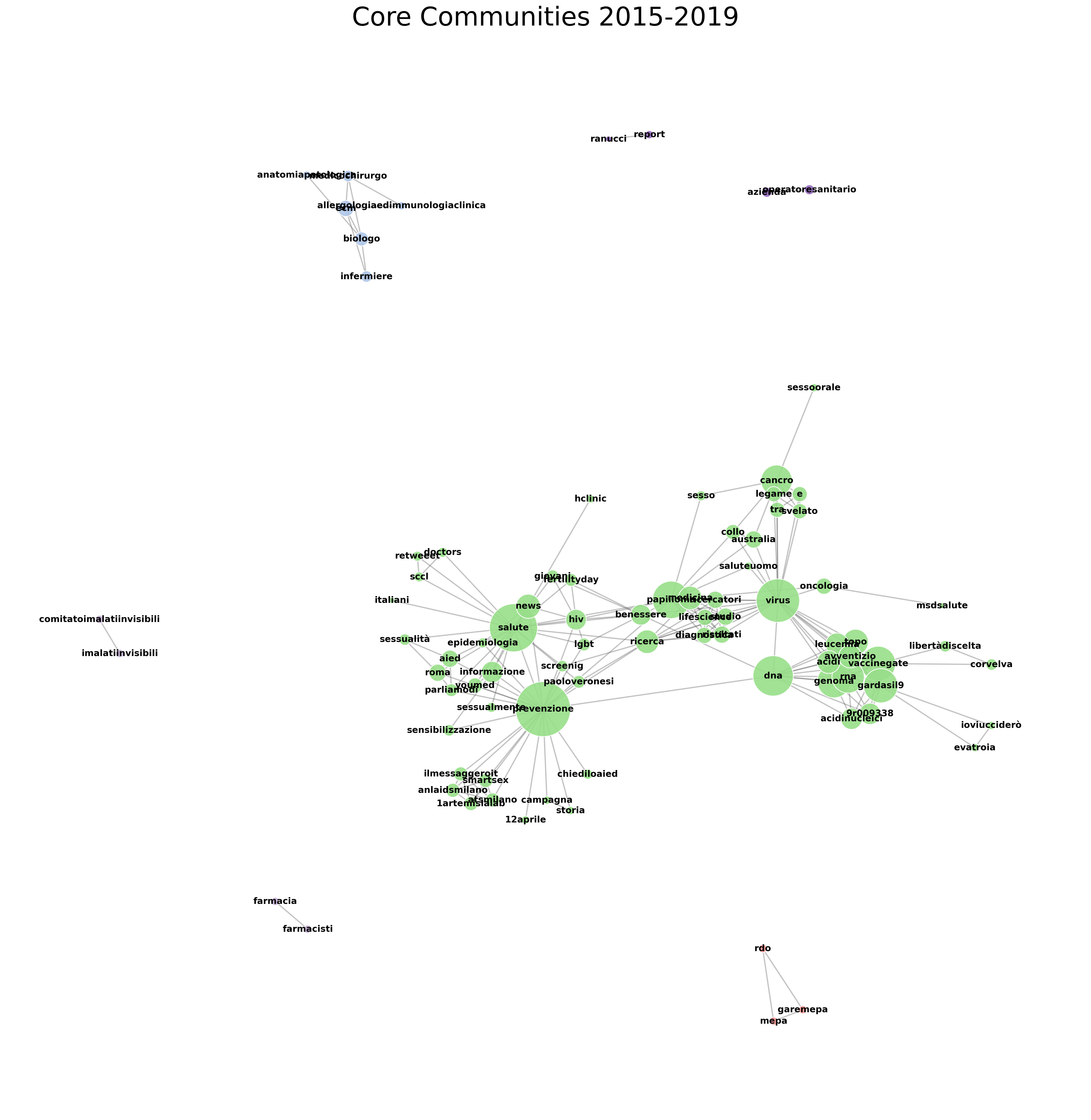}
    \caption{Core network (2015--2019).}
  \end{subfigure}
  \caption{Hashtag co-occurrence networks for 2015--2019 (coverage vs.\ core).}
  \label{fig:networks-2015-2019}
\end{figure*}

\begin{figure*}[!t]
  \centering
  \begin{subfigure}[t]{0.49\textwidth}
    \centering
    \includegraphics[width=\linewidth]{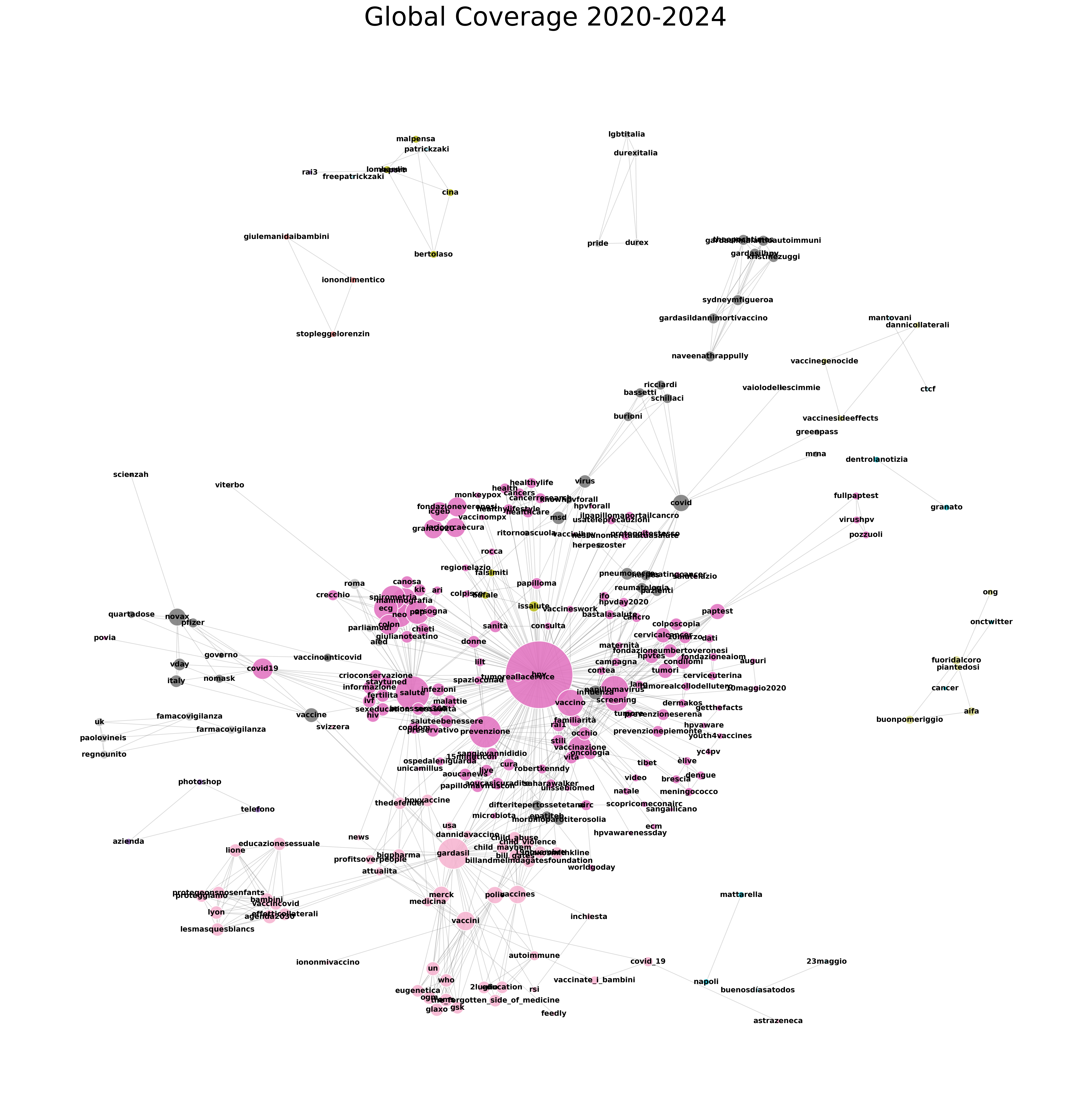}
    \caption{Coverage network (2020--2024).}
  \end{subfigure}\hfill
  \begin{subfigure}[t]{0.49\textwidth}
    \centering
    \includegraphics[width=\linewidth]{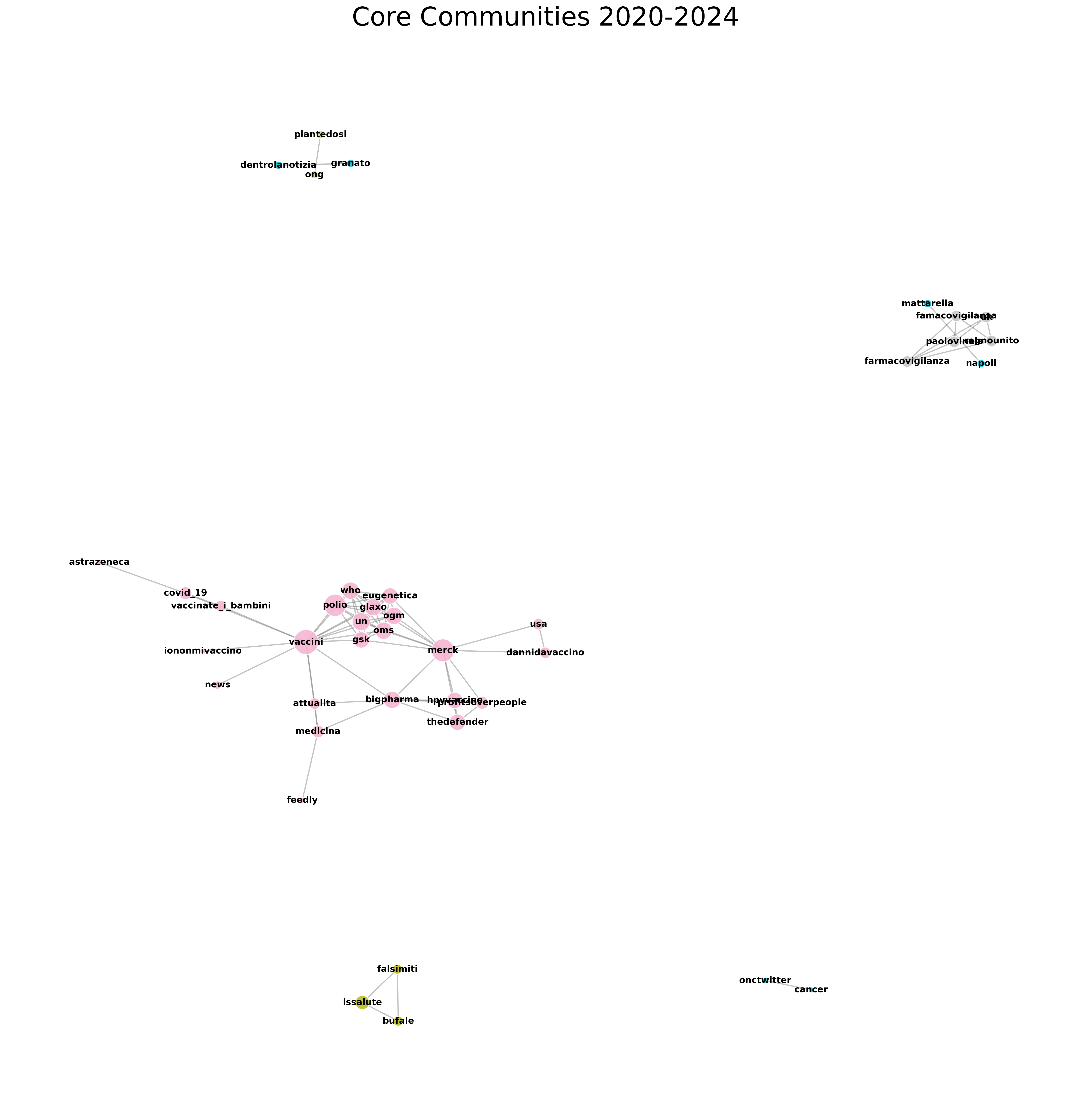}
    \caption{Core network (2020--2024).}
  \end{subfigure}
  \caption{Hashtag co-occurrence networks for 2020--2024 (coverage vs.\ core).}
  \label{fig:networks-2020-2024}
\end{figure*}

Figures \ref{fig:networks-2010-2014} and \ref{fig:networks-2020-2024} display the two layers as network graphs. The graphs show how the discourse expanded dramatically particularly between the first two epochs. The coverage layer, which represents the full conversational landscape, more than doubled in size in the epoch 2015-2019, growing from 150 nodes to 349 nodes, whereas the core layer grew from 46 to 86 nodes, indicating that more hashtags achieved the structural robustness required to be considered part of the core discourse. At the same time, while the network expanded, its density decreased, reflecting a more fragmented and specialized discourse where different communities become increasingly distinct. In coverage, the discourse is organized around the HPV anchor, as evidenced by the very high connectivity of the \#hpv node compared to other tags. Around it, stable health frames can be observed such as screening/prevention/cancer and medical/vaccine vocabulary (e.g., \#screening, \#prevenzione, \#cancro, \#tumore, \#paptest, \#vaccini, \#medicina, and \#gardasil). This suggests that early discourse is largely public-health informational and HPV is framed as prevention/cancer risk with a supporting vaccine thread. 
In contrast, the core is much smaller and notably does not contain \#hpv (nor edges involving \#hpv). This confirms that the core network foregrounds a tight prevention/screening/cancer backbone showing that the clinical/preventive narrative family is the most robust structure.

The second epoch (2015–2019) reveals the rise of narrative coalitions. While the prevention backbone persists (\#prevenzione, \#salute), a highly robust skepticism coalition has moved into the core. This community includes terms like \#vaccinegate, \#dna, \#genoma, and \#rna. The presence of these terms in the core indicates they are no longer fringe noise but have formed a structurally dense and persistent narrative coalition. The coverage layer captures the external political and social pressure. We see \#lorenzin (referring to the Italian mandatory vaccination law), \#novax, and pharmaceutical actors like \#merck. These terms are highly active but remain in the coverage layer, suggesting they are widely distributed across conversations rather than being anchored in a single tight thematic core.

The analysis of the final epoch (2020–2024) reveals a profound structural and thematic shift that characterizes the maturation stage of information disorder in the Italian HPV discourse. In this period, the network dynamics reflect a discourse deeply influenced by the post-COVID-19 environment and a consolidation of skeptical narratives. While the 2015–2019 epoch marked a peak in discursive expansion, 2020–2024 shows a consolidation. The coverage layer contains 268 nodes, slightly lower than the previous peak (349) but maintaining a high level of connectivity. The most striking result of this epoch is the content of the core layer. Unlike earlier periods where the core was dominated by institutional prevention, the 2020–2024 core has become the site of a highly robust, internationalized skepticism coalition. This is especially visible in the core community that features high-frequency hashtags such as \#bigpharma, \#eugenetica (eugenics), \#ogm, and \#who/\#oms (World Health Organization). The presence of \#covid\_19 within this same core cluster demonstrates that HPV discourse is no longer an isolated medical topic: it has been absorbed into a broader framework including other health topics. Moreover, the appearance of \#thedefender (associated with the Children’s Health Defense) and \#hpvvaccine alongside global pharmaceutical names like \#merck  and \#glaxo indicates that the Italian discourse is now tightly synchronized with international anti-vaccination narratives.

Another very significant result is the reversal of the 2010–2014 epoch, whereas many institutional and prevention-oriented terms have moved to the coverage layer in this later epoch. While hashtags like \#hpv, \#prevenzione, \#salute, and \#screening remain the most frequent in terms of total mentions (weighted degree), they are now part of the long tail in relation to the highly connected core of skepticism. The coverage layer also captures a broader range of general health services (\#mammografia, \#spirometria, \#ecg, \#paptest) suggesting that while the general public still discusses HPV in the context of screening, these conversations are structurally more fragmented compared to the tight, self-reinforcing core of reframed discourse. In other words, HPV-related discussion persists, but it co-exists with (and is partially recontextualised by) broader vaccine controversy, including anti-establishment and COVID-adjacent frames. 

From an information-disorder perspective, this doesn't merely show that new narratives appear, but that they alter the structural coupling between themes, whether by creating bridges that connect previously separate frames, or by forming parallel clusters that remain partly segregated yet reshape the overall discourse environment.

\section{Validation}
\label{sec:seeds}
To validate the methodology, we connected hashtag communities to higher-level information-disorder narratives by constructing a small seed lexicon of narrative-indicative hashtags for two frames: \emph{skeptical/anti-vaccine} and \emph{conspiracy}. The skeptical/anti-vaccine frame included posts expressing doubt or opposition to vaccination, such as safety/efficacy concerns and institutional distrust framed as skepticism while the conspiracy frame was used to identify posts alleging coordinated deception or malicious intent, e.g., Big Pharma plots, depopulation/sterilization claims, censorship narratives. The author manually annotated a small sub-set of randomly chosen posts ($\approx 10 $ per year); posts were flagged conservatively only when the narrative signal was unambiguous.
From this subset, we identified 86 seed posts (69 conspiratorial and 17 skeptical). These posts yielded a seed-hashtag lexicon comprising 52 conspiracy-associated hashtags, 5 skeptical hashtags, and 3 overlapping hashtags. This lexicon was then used as weak supervision to calculate community-level enrichment scores, indicating whether seed-linked content was overrepresented in a given community relative to its size. The list of seed hashtags is in Table \ref{tab:seed-hashtags} in the \hyperref[appendix]{Appendix} whereas Table \ref{tab:seed-examples} shows two examples of posts for both categories.

\begin{table}[t]
\centering
\footnotesize
\setlength{\tabcolsep}{3pt}
\begin{tabular}{l p{0.36\columnwidth} p{0.36\columnwidth}}
\hline
\textbf{Label} & \textbf{Italian post} & \textbf{English translation} \\
\hline
Skeptical &
VIDEO HPV: I GIOVANI CHIEDONO PIU' INFORMAZIONE SUL VACCINO &
HPV VIDEO: Young people ask for more information about the vaccine \\
Skeptical &
Tumori: vaccino hpv non convince mamme &
Cancer: the HPV vaccine does not convince mothers \\
Conspiracy &
Yes, \`e proprio un giornalista ad aver iniziato, con una serie di articoli scandalosi... per quanto riguarda la comunit\`a scientifica, no comment &
Yes, it was a journalist who started it, with a series of scandalous articles... as for the scientific community, no comment \\
Conspiracy &
non sono i giornali ad aver affossato certe teorie ma la stessa comunit\`a scientifica internazionale.. il caso wakefield \`e addirittura peggio &
It wasn't the newspapers that sank certain theories, but the international scientific community itself... the Wakefield case is even worse \\
\hline
\end{tabular}
\caption{Examples of posts flagged by the skeptical and conspiracy seed indicators, with English translations.}
\label{tab:seed-examples}
\end{table}

\begin{figure}[t]
  \centering
  \includegraphics[width=\columnwidth]{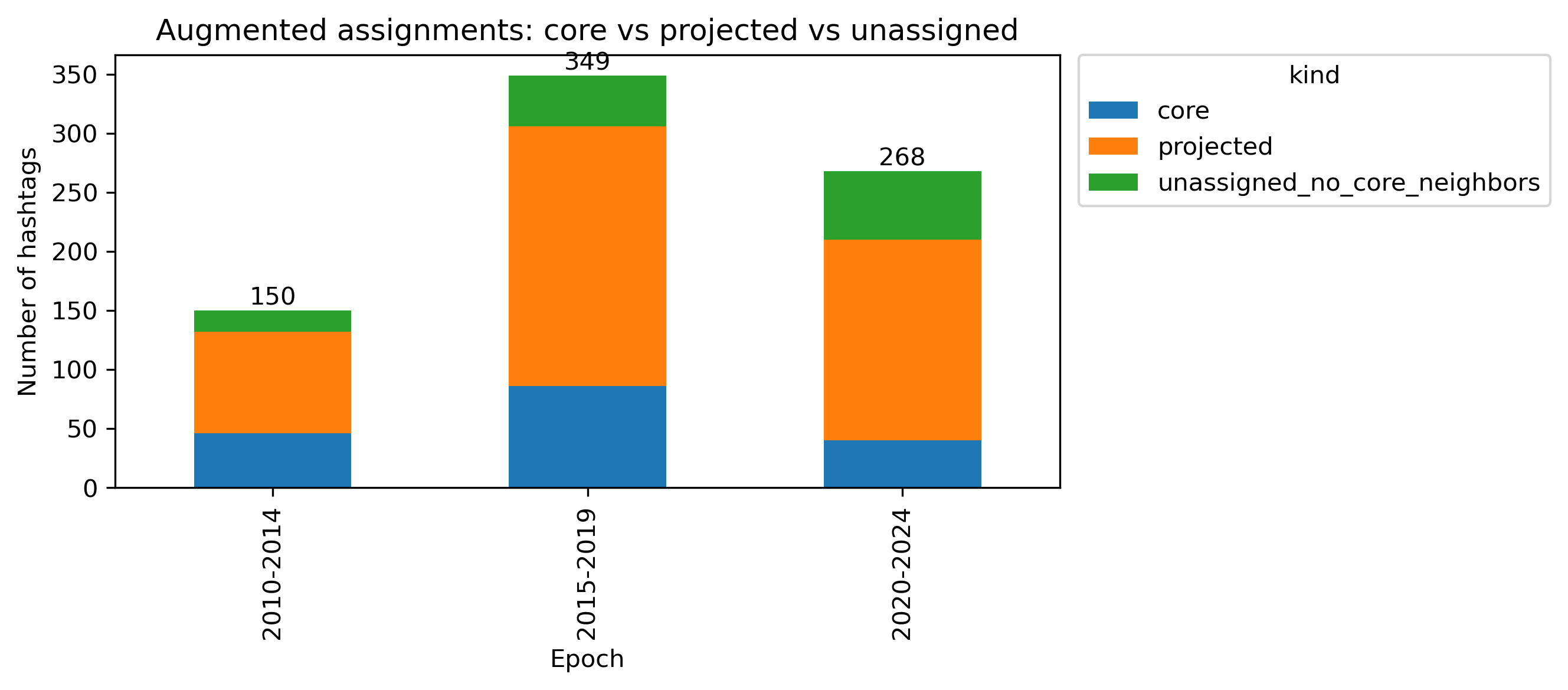}
  \caption{Augmented assignments by epoch.}
  \label{fig:augmented-assignments}
\end{figure}

For each epoch and community, the seed enrichment score was computed as the ratio between (i) the community’s share of all seed mentions and (ii) the community’s baseline share of total hashtag mentions. Communities with enrichment $>1$ are over-represented in the corresponding narrative seeds relative to their size, proving that the community contains a higher concentration of skeptical or conspiratorial content than the rest of the network. Figure ~\ref{fig:augmented-assignments} shows how in each epoch, only a minority of hashtags belong to the core backbone used for community detection, while a larger set of hashtags can be projected onto the core community structure based on their connectivity to core nodes. A third group remains unassigned, indicating peripheral or disconnected usage that cannot be reliably linked to a community. Figure ~\ref{fig:projection-support} in the \hyperref[appendix]{Appendix} reports the distribution of the projection support score (the summed weight of edges linking a projected hashtag to its assigned core community).

\begin{figure*}[!t]
  \centering
  \begin{subfigure}[t]{0.49\textwidth}
    \centering
    \includegraphics[width=\linewidth]{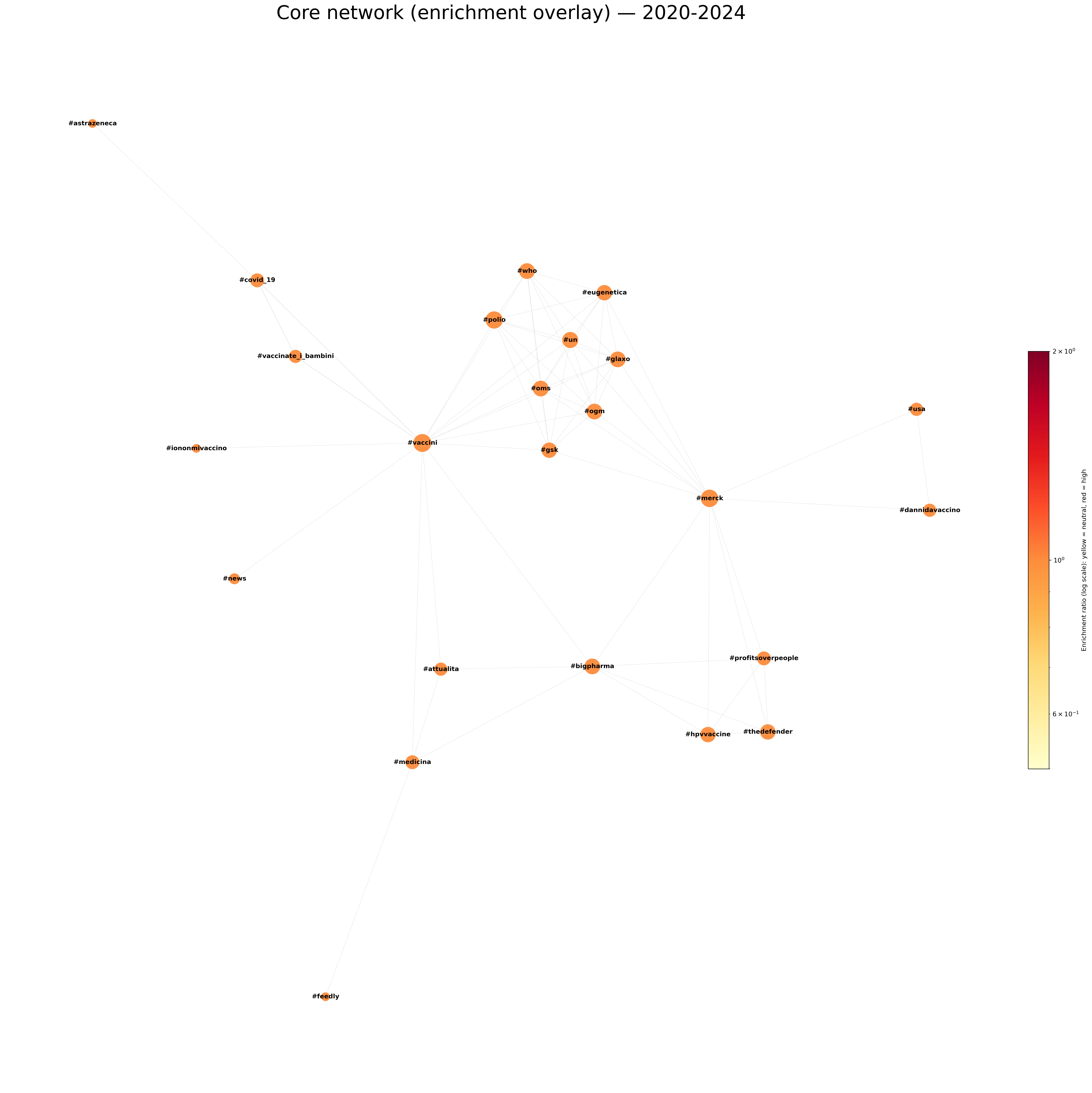}
    \caption{Core network (backbone).}
    \label{fig:core-enrich-2020-2024}
  \end{subfigure}\hfill
  \begin{subfigure}[t]{0.49\textwidth}
    \centering
    \includegraphics[width=\linewidth]{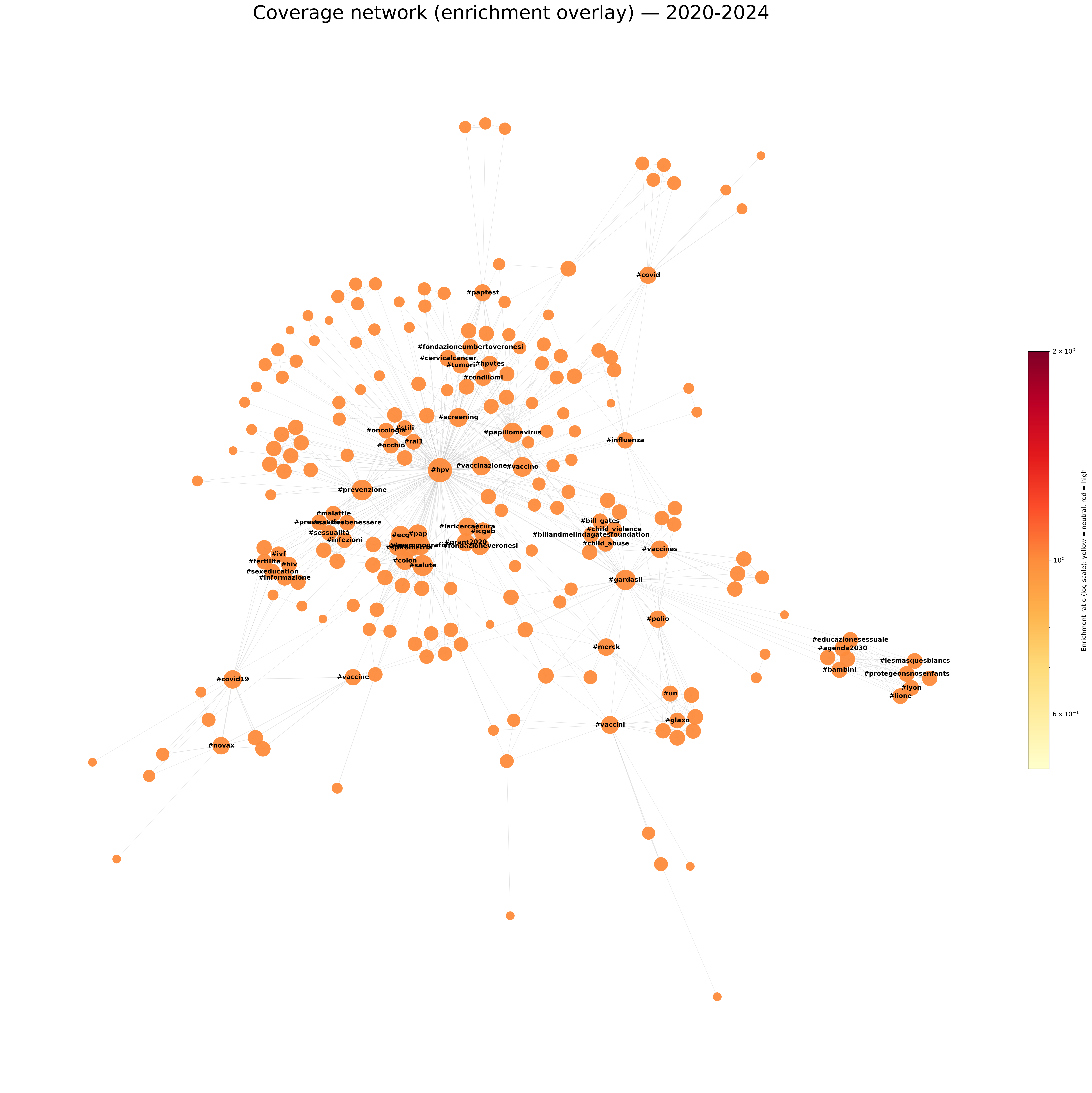}
    \caption{Coverage network (LCC).}
    \label{fig:cov-enrich-2020-2024}
  \end{subfigure}
  \caption{{Core vs coverage enrichment overlay (2020--2024).}}
  \label{fig:core-vs-coverage-enrichment-2020-2024}
\end{figure*}

As an example of the core vs coverage enrichment overlay, Figure \ref{fig:core-vs-coverage-enrichment-2020-2024} displays the enriched graph for the third epoch (2020–2024). Nodes are coloured by community-level seed enrichment (yellow = baseline, red = high enrichment; log-scaled). The core network (left) shows the sparsified backbone used for community detection; the coverage network (right; largest connected component) retains peripheral hashtags and bridging edges. High enrichment appears in a small number of network communities rather than across the network as a whole. This indicates that information-disorder markers cluster into coherent sub-discourses with consistent co-occurring hashtags, instead of being randomly or uniformly scattered throughout the HPV/vaccine conversation. Importantly, because enrichment is defined relative to community size, concentrated high values reflect disproportionate over-representation of seed markers within particular sub-narratives.

\section{Discussion}
The 14-year longitudinal analysis (2010–2024) provides empirical evidence for the lifecycle of HPV vaccine discourse in Italian X through narrative coalitions. In the three epochs, clear core clusters shifts can be identified: institutional discourse forms the core in 2010-2014 and skepticism is sparse and located in the long tail. In the second epoch (2015–2019), narrative coalitions around controversy and skepticism mature into a core cluster that competes with the prevention backbone. Finally, in the last epoch (2020–2024), the \emph{information disorder} coalition solidifies into a dominant core. The HPV discourse is no longer about the vaccine but about reframing health as a political and conspiratorial struggle against global institutions. 

Methodologically, the dual-layer approach proved effective for detection of early information disorder signals. The coverage networks ensured that peripheral and bridging material was not excluded, while the core networks continue to identify the most robust associations that were least sensitive to low-frequency noise. The fact that in 2020–2024, the core backbone contains almost no links between different communities suggests that late-period discourse is more segmented into distinct frames. 

Finally, by using manually labelled seed posts (skeptical/conspiratorial) to validate the projection, the method effectively treats hashtags as cultural products, providing a sociolinguistic lens onto how hashtags can hijack narratives. This provides a clear empirical basis for interpreting the results as a form of conceptual change in which stable health concepts persist but become embedded in a reconfigured and more polarised information environment.
The findings demonstrate that information disorder is also matter of structural shifts in connectivity. In turn, this highlights the importance of the methodology introduced here: it captured that the projected influence of institutional discourse still exists but has lost its central structural dominance.


\section{Conclusion}
\label{sec:conclusions}
This paper examined the temporal evolution of Italian discourse about the HPV vaccine on X through a graph-based analysis of hashtag co-occurrence. By modelling hashtags as cultural objects that mark discourse and tracking their co-occurrence structure across three epochs (2010-2014, 2015-2019, 2020-2024), cross-epoch community matching, and augmented projection, we show that the HPV conversation undergoes a clear process of conceptual reconfiguration. 

Beyond the HPV case, the proposed methodology provides a transparent and reusable framework for studying conceptual change in online discourse and for linking network communities to information-disorder narratives via seed enrichment. The innovation of the method lies in its attempt to bridge the gap between Discourse Studies and Network Science, specifically by addressing the loss of signal problem in longitudinal online discourse studies. 

The Core–Coverage projection effectively solves the long tail issue which typically results in graphs structurally biased towards mainstream/dominant discourse. By capturing the coverage layer and then projecting those fringe nodes into the core, this framework creates a computational zoom lens that allows to see when sparse, low-frequency hashtags are reframing attempts that attach themselves to established narrative backbones. In this way, the analysis is moved beyond what is being said to how the structure of the argument matures. 

\section{Limitations}
While this study provide several insights, limitations must be acknowledged. First, the dataset is subject to the X’s API access constraints and policy requirements and observed temporal differences may partially reflect changes in platform usage and data availability rather than discourse alone. Second, hashtag-based analysis captures a salient but incomplete layer of meaning-making: many tweets contain no hashtags, hashtags are used strategically (for visibility, irony, or audience targeting), and co-occurrence networks do not directly encode stance, intent, or factual accuracy. Therefore, the reliance on hashtags as primary signals may overlook important discourse without hashtags. Third, community structure depends on modelling choices (epoch boundaries, edge thresholds, top-$k$ pruning, $k$-core filtering, and community detection settings). While we mitigate this by reporting core vs.\ coverage results and by using augmented projection with explicit support scores, different parameterisations may yield alternative granularities of communities. 

The seed-based enrichment approach provides weak supervision for narrative-family mapping but is sensitive to seed selection and ambiguity (e.g., topic anchors that co-occur with narrative markers), therefore the seed-based validation may introduce bias depending on selection criteria. Enrichment serves as a guide for qualitative validation rather than a definitive classifier of conspiracy or skepticism.  Additionally, while the method is generalizable in principle, further evaluation on other domains or datasets would strengthen the claims.
Future work should extend the analysis with larger corpora and validate narrative-family mapping through systematic annotation and inter-coder agreement, enabling stronger claims about misinformation, disinformation, and malinformation dynamics.

 
\section{References}
\bibliographystyle{lrec2026-natbib}
\bibliography{latex/references}

\appendix
\section{Appendix}
\label{appendix}

\begin{table}[t]
\centering
\footnotesize
\setlength{\tabcolsep}{4pt}
\begin{tabular}{l r}
\hline
\textbf{Statistic} & \textbf{Value} \\
\hline
Time span (UTC) & 2010-01-02 to 2024-12-30 \\
Duration & 5{,}476 days ($\approx$15.0 years) \\
Tweets (rows) & 4{,}895 \\
Unique users (user IDs) & 1{,}910 \\
Unique usernames & 2{,}252 \\
Tweets with $\ge$1 hashtag & 904 (18.47\%) \\
Unique hashtags (observed) & 754 \\
\hline
Likes (sum / median / max) & 191{,}523 / 0 / 28{,}662 \\
Replies (sum / median / max) & 25{,}864 / 0 / 1{,}314 \\
Retweets (sum / median / max) & 36{,}562 / 0 / 2{,}261 \\
Quotes (sum / median / max) & 5{,}215 / 0 / 425 \\
\hline
\end{tabular}
\caption{Descriptive statistics for the Italian-language Twitter/X corpus. Hashtag presence is computed across all hashtag fields (including quoted/reply tweet objects when available).}
\label{tab:dataset-stats}
\end{table}

\begin{table}[t]
\centering
\footnotesize
\setlength{\tabcolsep}{3pt}
\begin{tabular}{l p{0.72\columnwidth}}
\hline
\textbf{Seed set} & \textbf{Unique seed hashtags} \\
\hline
Skeptical &
\#feedly, \#gardasil, \#hpv, \#medicina, \#papilloma \\
Conspiracy &
\#algoritmo, \#antivax, \#assassinidistato, \#aulascienze, \#bigpharma, \#burioni, \#business, \#butac, \#casefarmaceutiche, \#censura, \#checoincidenza, \#cicap, \#crimini, \#facebook, \#gardasil, \#giulemanidaibambini, \#giulemanidaigiovani, \#hpv, \#ionondimentico, \#iovaccini, \#iovaccino, \#laverit\`a, \#libertadiscelta, \#lorenzin, \#mafiamedica, \#malattiesessuali, \#medicina, \#merck, \#nonmiafiglia, \#notizie, \#novax, \#papillomavirus, \#portatore, \#pseudomedicina, \#salute, \#sanita, \#sano, \#scienza, \#scuola, \#senzacategoria, \#speriemtazione, \#standingupforscience, \#stopleggelorenzin, \#strinic, \#teamvaxitalia, \#ultimora, \#uominiedonne, \#uomo, \#vaccinarsi, \#vaccinazione, \#vaccinazioni, \#vaccini \\
Overlap  &
\#hpv, \#gardasil, \#medicina \\
\hline
\end{tabular}
\caption{Seed hashtag lexicon used to flag skeptical and conspiracy narratives (hashtags normalised to lowercase).}
\label{tab:seed-hashtags}
\end{table}

\begin{figure}[t]
  \centering
  \includegraphics[width=\columnwidth]{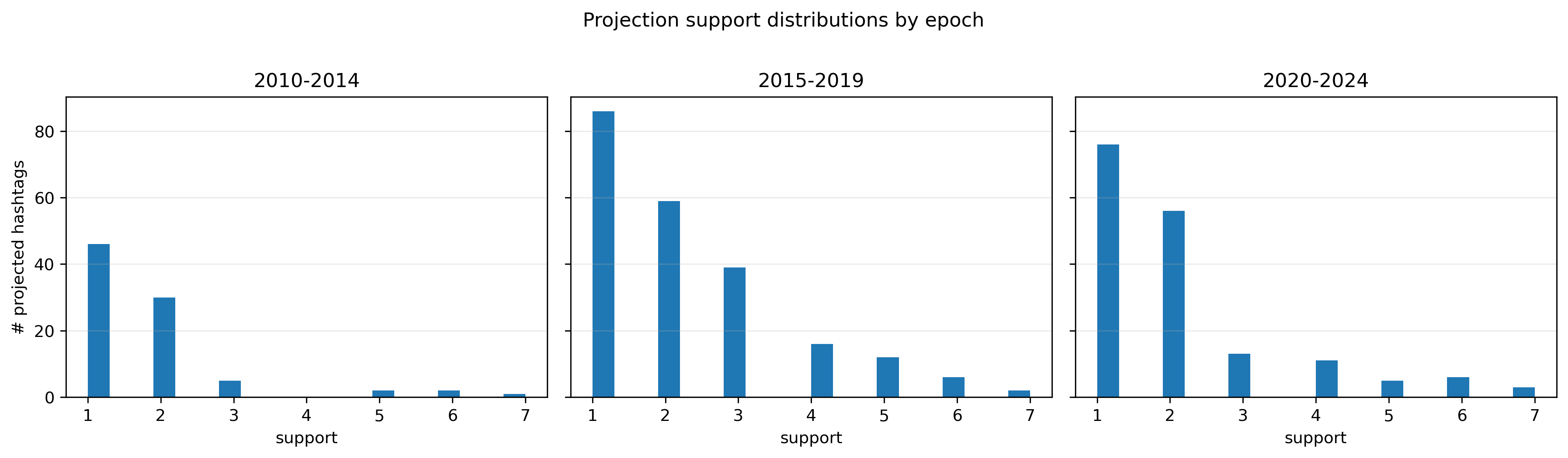}
  \caption{Distribution of projection support by epoch (sum of edge weights from a projected hashtag to its assigned core community).}
  \label{fig:projection-support}
\end{figure}

\end{document}